\documentclass[11pt,a4paper]{article}
\usepackage[hyperref]{acl2021}
\usepackage{times}
\usepackage{latexsym}

\usepackage{microtype}
\usepackage{hyperref}
\usepackage{url}
\usepackage{multirow}
\usepackage{graphicx}
\usepackage{tabularx}
\usepackage{varwidth}
\usepackage{algpseudocode}
\usepackage{algorithm}
\usepackage[english]{babel}
\usepackage[autostyle]{csquotes}
\usepackage{amsmath,amssymb}
\DeclareMathOperator{\E}{\mathbb{E}}
\DeclareMathOperator*{\argmax}{arg\,max}
\algnewcommand\algorithmicforeach{\textbf{for each}}
\algdef{S}[FOR]{ForEach}[1]{\algorithmicforeach\ #1\ \algorithmicdo}

\makeatletter
\newcommand{\thickhline}{%
    \noalign {\ifnum 0=`}\fi \hrule height 1pt
    \futurelet \reserved@a \@xhline
}
\newcolumntype{"}{@{\hskip\tabcolsep\vrule width 1pt\hskip\tabcolsep}}
\makeatother

\makeatletter
\newcommand{\StatexIndent}[1][3]{%
  \setlength\@tempdima{\algorithmicindent}%
  \Statex\hskip\dimexpr#1\@tempdima\relax}
\algdef{S}[WHILE]{WhileNoDo}[1]{\algorithmicwhile\ #1}%
\makeatother

\aclfinalcopy 

\title{$\textrm{WeaSuL}^{\pi}$: Weakly Supervised Dialogue Policy Learning: Reward Estimation for Multi-turn Dialogue}

\author{Anant Khandelwal\\
  India Machine Learning\\
  Amazon \\
  \texttt{anantkha@amazon.com} \\}

\date{}

\begin{document}
\maketitle
\begin{abstract}
An intelligent dialogue system in a multi-turn setting should not only generate the responses which are of good quality, but it should also generate the responses which can lead to long-term success of the dialogue. Although, the current approaches improved the response quality, but they over-look the training signals present in the dialogue data. We can leverage these signals to generate the weakly supervised training data for learning dialog policy and reward estimator, and make the policy take actions (generates responses) which can foresee the future direction for a successful (rewarding) conversation. We simulate the dialogue between an agent and a user (modelled similar to an agent with supervised learning objective) to interact with each other. The agent uses dynamic blocking to generate ranked diverse responses and exploration-exploitation to select among the Top-K responses. Each simulated state-action pair is evaluated (works as a weak annotation) with three quality modules: Semantic Relevant, Semantic Coherence and Consistent Flow. Empirical studies with two benchmarks indicate that our model can significantly out-perform the response quality and lead to a successful conversation on both automatic evaluation and human judgement.
\end{abstract}

\section{Introduction}
Dialog policy for multi-turn dialogue decides the next best action to take on the environment so as to complete the conversation based on various success criteria. Reinforcement learning can help to learn such a policy where the environment can be users (human or model) and the policy takes action on the environment from which it gets a reward signal \cite{fatemi2016policy, peng2017composite, chen2017agent, yarats2018hierarchical, lei2018sequicity, he2018decoupling, su2018discriminative}.

Learning a dialogue policy using reinforcement learning can be challenging with humans users, since it requires a large set of samples with a reward to train. Since there are a lot of previous works on neural response generation \cite{gu2020dialogbert, zhao2020learning, zhang2019recosa, xing2018hierarchical, serban2016building} we can model the users also, using any of these encoder-decoder architectures. This helps to simulate the conversations between the simulated user and the agent (policy model) replying to each other \cite{zhao2016towards, dhingra2016towards, shah2018bootstrapping}. Reward signal for policy learning can be as simple as the small constant negative reward at each turn and a large reward at the end (if the goal completes) to encourage shorter conversations \cite{takanobu2019guided}.

However, reward estimation for dialogue is challenging, the small constant negative reward at each turn may lead to ending the conversation prematurely. Instead of handcrafting the reward at the end based on success or failure, it is more useful if we can evaluate reward at every turn to guide the policy to dynamically change actions as per the need for the user and end the conversation naturally. With the growing complexity of the system across different topics, it is required to build a more sophisticated reward function to avoid manual intervention for accounting different factors towards conversation success.

In this work, we proposed a novel model for contextual response generation in multi-turn dialogue. The model includes the turn-level reward estimator, which combines the weak supervision signals obtained from three basic modules 1) Semantic Coherence, 2) Consistent Flow, 3) Semantic Relevance. These modules are learned jointly with the response generation model with the counterfactual examples obtained from negative sampling. Leveraging the weak supervision signals obtained from these models, we further update the reward estimator and dialog policy jointly in an alternative way, thus improving each other.

Our proposed approach integrates semantic understanding of utterances using encoder-decoder systems with the power of Reinforcement Learning (RL) to optimize long-term success. We test the proposed approach with two benchmarks: DailyDialog \cite{li2017dailydialog} and PersonaChat \cite{zhang2018personalizing}. Experimental results demonstrate on both datasets indicate that our model can significantly outperform state-of-the-art generation models in terms of both automatic evaluation and human judgment.

\section{Related Work}
Open-domain dialogue in a multi-turn setting has been widely explored with different encoder-decoder architectures \cite{gu2020dialogbert, feng2021multi, kottur2017exploring, li2016deep, shah2018bootstrapping, shang2015neural, vinyals2015neural, wu2019self, zhao2020learning, zhong2019affect}. The basic encoder-decoder architectures like Seq-to-Seq models have been widely extended and modified to generate the generic responses, context modelling and grounding by persona/emotion/knowledge \cite{li2015diversity, xing2017topic, serban2016building, xing2018hierarchical, zhang2019recosa, zhang2018personalizing, zhou2018emotional, dinan2018wizard}.

The dialogue literature widely applies reinforcement learning, including the recent ones based on deep architectures \cite{takanobu2019guided, takanobu2020multi, li2020guided, takanobu2020multi, li2020guided, gordon2020learning, gordon2020show}. But these task-oriented RL dialogue systems often model the dialogue with limited parameters and assumptions specific to the dataset, targeted for that task. The dataset includes hand-built templates with state, action and reward signals designed by humans for each new domain making this setting difficult for extending these to open domain dialogue systems.

Our goal in this work is to integrate the state-of-the-art encoder-decoder architectures like in \citet{gu2020dialogbert, zhao2020learning, csaky2020gutenberg} and reinforcement learning paradigms to efficiently learn the dialogue policy optimized for long-term success in the multi-turn dialogue scenarios. We are recently inspired by the works in \citet{takanobu2019guided, li2020guided, li2016deep} to jointly learn the reward function and dialogue policy, and reduce the effort and cost for manual labelling the conversations for building the reward  model. Specifically, we leverage the weak supervision inspired from \citet{chang2021jointly, chang2021neural} to generate the labelled dataset to facilitate this joint learning and building reward estimation model.

\section{Approach}
We represent dialog sessions $\mathcal{D} = \{\tau_1, \tau_2, \tau_3,.......\tau_n\}$ where each dialog session $\tau$ represents the trajectory of state-action pairs as $\{s_0^{u}, a_0^{u}, s_0, a_0, s_1^{u}, a_1^{u}, s_1, a_1, .....\}$. The user in our case is a simulator which utters a response $a^{u}$ given the state $s^{u}$ denoted as $\mu(a^{u}, e^{u} | s^{u} )$ where $e^u$ denotes the binary signal indicating the end of a dialog session, in that case 
the response $a^u$ is empty. The dialog policy $\pi_{\theta}(a | s)$ decides the action $a$ according to the current state $s$ after the agent interacts with the user simulator $\mu$. At each time, the state given to the either dialog party is updated after recording the action uttered by the other party. The reward estimator $f$ evaluates the quality of response/action uttered by the dialog policy $\pi$. The dialog policy $\pi$ is based on the BERT \cite{devlin-etal-2019-bert} encoder-decoder model and the reward function $f$ is the MLP model parameterized by $\theta$ and $\omega$ respectively. We have modeled the user simulator exactly in the same way as the agent but trained only using supervised learning objective. 

In the subsequent section, we will introduce the components action, state, policy, quality modules and reward estimator. Further, sections explain the setup we have used for weakly supervised learning and, finally, the experimental results.

\subsection{Action}
\begin{figure}
    \centering
    \includegraphics[width=\linewidth]{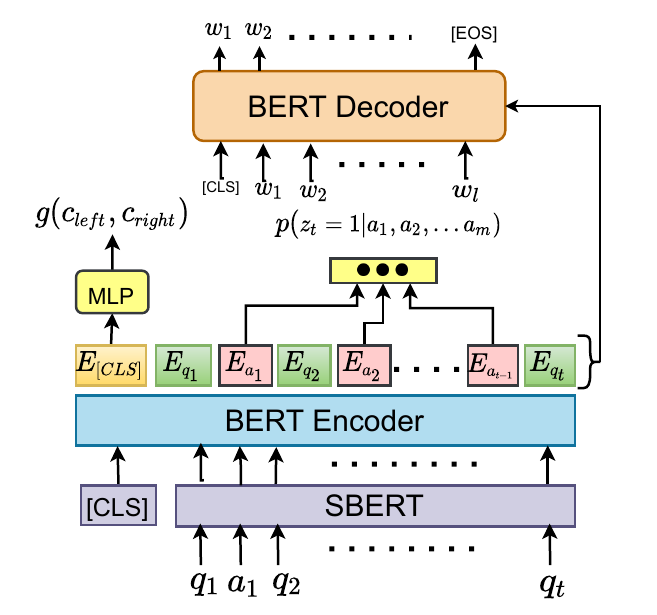}
    \caption{BERT based Encoder-Decoder with Semantic Coherence and Relevance. Similarly, Consistent Flow loss is also calculated using encoder.}
    \label{fig:my_label}
\end{figure}

An action $a$ is the dialogue utterance generated by the encoder-decoder model as shown in Figure \ref{fig:my_label}. The model takes as input the context history (state), and outputs the probability distribution over a set of possible actions denoted as $\pi_{\theta}(a | s)$ parameterized by $\theta$. The user simulator generates the action $a^{u}$, policy generates the action $a$, and the input state for the agent and the user is $s$ and $s^u$ respectively.

\subsection{State}
The state is the past conversation history between an agent and a user denoted as, $s_t = \{q_1, a_1, q_2, a_2, q_3, a_3,.....,q_{t}\}$. The state for an agent and a user are differently denoted as $s$ and $s^{u}$ respectively. Let's say the agent utterances are denoted by $a$'s, then state, $s = s_t$ and the agent utters $a_t$. Similarly, the user state $s_t^u = \{q_1, a_1, q_2, a_2, q_3, a_3,.....,q_{t}, a_t\}$ and the user utters $q_{t+1}$. Each of the utterances is mapped to a fixed-length sentence vector using SBERT \cite{reimers2019sentence}. 

\subsection{Dialogue Policy}
The dialogue policy takes the form of a BERT based encoder-decoder ( i.e. $\pi_{\theta}(a | s)$ ) \cite{gu2020dialogbert} as shown in Figure \ref{fig:my_label}. Similar to \citet{xu2020learning}, we have used the BERT based encoder and transformer decoder, but instead of feeding the utterance at word level, we instead fed the utterance representation (obtained from SBERT) into the encoder. The encoder takes as input the previous context history as $s_t$ and output the response $a_t$ at the output of the decoder.

\subsection{User Simulator}
We have modelled the user simulator in exactly the same way as the BERT based encoder-decoder shown in Figure \ref{fig:my_label}. However, the user simulator is trained only (with supervised learning objective) for utterances in dialog corpus and predicting user response \cite{gu2020dialogbert}. 

\subsection{Conversation Quality Modules}
We calculate the reward for each state-action pair (see Section. \ref{weasul}) and use this signal to train the dialogue policy so that it can avoid reaching bad states so as to reach the successful end of the conversation between a user and an agent. We have leveraged the signals from three basic modules, namely, Semantic Coherence, Consistent Flow and Semantic Relevance (which are jointly learned with the dialogue policy). For each of the three modules, the data for the positive class is obtained from the source corpus while for the negative class it has been generated dynamically during training. We describe each of the three modules in the following sections.

\subsubsection{Semantic Relevance}
We need to filter out the utterances generated with high confidence by the dialog policy but are semantically irrelevant to the previous context. To quantify such a characteristic, we modeled the general response relevance prediction task which utilizes the sequential relationship of the dialog data fed to the encoder side of BERT encoder-decoder framework. Since, the task of semantic relevance is to match the two sequences of conversation, so instead of matching the context and response, we have measured the relevance of two fragments of dialogue session.

Specifically, given a context $c = \{q_1, a_1, q_2, a_2, . . . . . q_m\}$, we randomly split $c$ into two consecutive pieces $c_{left} = \{q_1, a_1, q_2, a_2, ....q_t, a_t\}$ and $c_{right} = \{q_{t+1}, a_{t+1},.....q_{m}\}$. Similar to \citet{xu2020learning}, we replaced the left or right part with the sampled piece from the corpus. Also, we additionally generate the negative samples by internal shuffling in the left or right part. The whole model is trained like a classifier with corresponding labels $y_{sr} \in \{0, 1\}$. Since the individual utterances are fed after obtaining their vector representation, the aggregated representation of two pieces is represented by $E_{CLS}^{sr}$ over which the non-linear transformation is applied, the score for semantic relevance is given by $g(c_{\textrm{left}}, c_{\textrm{right}})$, and similar to \citet{xu2020learning}, it has been trained using the binary cross-entropy loss as:
\begin{multline}
    L_{sr} = -y_{sr} \log(g(c_{\textrm{left}}, c_{\textrm{right}})) \\
- (1 - y_{sr}) \log(1 - g(c_{\textrm{left}}, c_{\textrm{right}}))
\end{multline}
\subsubsection{Semantic Coherence}
The response generated should be rewarded only if it is coherent despite having adequate content. This makes the model to generate the coherent responses while avoiding the incoherent ones. Specifically, given a context $c = \{q_1, a_1, q_2, a_2, . . . . . q_m\}$, we randomly select any of the agent response at time $t$, denoted as $a_t$, and replace it with any random utterance from the corpus. We also generate the incoherent samples by internal shuffling of bi-grams. The incoherent utterance is labelled as $y_{t}^{coh} = 0$ and coherent samples as $y_t^{coh} = 1$. The semantic coherence model is also trained like a classifier for each of the utterance representations obtained at the output of BERT encoder as shown in Figure \ref{fig:my_label}. The probability of the $t$-th utterance being incoherent is given as:
\begin{multline} 
p(z_t = 1| a_1,.., a_t) = softmax(W_{coh}E_{a_t} + b_{coh}) \\
    = \frac{exp(W_{coh}E_{a_t} + b_{coh})}{\sum_{l=1}^{m} exp(W_{coh}E_{a_l} + b_{coh})}
\end{multline}
and the loss function is given as:
\begin{equation}
    L_{coh} = -\sum_{t=1}^{m} z_t \log p(z_t = 1 | a_1, a_2 ..... a_m)
\end{equation}

\subsubsection{Consistent Flow}
We want the agent to continuously add the information to keep the conversation going in the forward direction. To determine the flowing conversation, we take the cosine similarity between the last two agent utterances denoted as 
$E_{a_{i - 1}}$ and $E_{a_{i}}$ denoted as $g(a_{i - 1}, a_{i})$, and we measure the similarity with randomly sampled utterance $v$ in place of $a_{i-1}$ given as $g(a_{i-1}, v)$. We would like $g(a_{i-1}, a_{i})$ to be larger than $g(a_{i-1}, v)$ by at least a margin $\Delta$ and define the learning objective as a hing loss function:
\begin{multline}
        L_{cf} = max\{0, \Delta - g(a_{i-1}, a_{i}) + g(a_{i - 1}, v)\}
        \vspace{-40mm}
\end{multline}
\subsection{Joint Training of Agent and Reward Modules}
\label{tpj}
To initialize the parameters of agent and reward modules $\mathcal{M} = $\{Semantic Relevance, Semantic Coherence, Consistent Flow\}, we used the supervised learning objective since all the state-action pairs obtained from the pre-training corpus are the ground-truth and can be used as close approximation for further fine-tuning on other dialog corpus. We used the pre-training corpus $\mathcal{P}$ as Gutenberg dialog corpus \cite{csaky2020gutenberg}. Since the agent model in our case is based on BERT encoder-decoder parameterized by $\theta$ similar to \citet{gu2020dialogbert}, the probability of generating agent's response $\textbf{a}$ is given as:
\begin{equation}
    p_{\theta}(\textbf{a} | \textbf{s} ) = \prod_{j=1}^{N} p_{\theta}(a_j | a_{<j}, \textbf{s}),
\end{equation}
where $a_j$ is the j-th word generated at the output of decoder and $\textbf{s}$ is the whole context history utterances fed to the encoder and $N$ is the maximum sequence length of decoder. The loss function for generating agent response $\textbf{a}$ is given as:
\begin{equation}
    L_a = J(\theta) = -\sum_{i=1}^{N} \log p_{\theta}(a_j|a_{<j}, \textbf{s})
\end{equation}
The joint loss function is defined as:
\begin{equation}
    L_{full} = L_{a} + \alpha* ( L_{sr} + L_{coh} + L_{cf} )
\end{equation}
The policy $\pi_{\theta}$ is also parameterized by $\theta$, and the probability of action $a$ is given by $\pi_{\theta}(a | s)$ similar to $p_{\theta}(a | s)$, since the probability distribution is learned only from $(s, a)$ pairs obtained from the corpus with human demonstrations. It is a good approximation to initialize the parameters of policy $\pi_{\theta}(a | s)$ with parameters of $p_{\theta}(a | s)$. Furthermore, we update the policy $\pi_{\theta}$ (Step 13 in the Algorithm. \ref{ref:algo}) to avoid actions $a$ which do not lead to rewarding conversations.
\subsection{Dialogue Simulation between Agent and User}
\label{diagsim}
We setup simulation between virtual agent and user, and let them take turns talking to each other. The simulation is started with a starter utterance obtained from the dialog samples $D_H$ (Step 5 of Algorithm \ref{ref:algo}) and fed to the agent, it then encodes the utterance and generates the response $a$, the state $s^{u}$ is then updated with previous history and fed to the user model to obtain the next response $a^u$. The response $a^u$ is appended to $s^u$ to obtain the updated state $s$. Similarly, the process is repeated until one of the following conditions occurs after a few number of turns\footnote{The number of turns after these rules applied is average number of turns in the corpus}: a) When agent starts to produce dull responses like \enquote{I don't know} \footnote{Used simple rule matching method with 9 phrases collected from the corpus, instead of having false positives and negatives this works well in practice.}. b) When agent starts to generate repetitive response consecutively \footnote{If by rule two consecutive utterances matched more than 80\% it is said to be repetitive.} c) Or, the conversation achieved the maximum number of turns handled by agent and user models.\footnote{The maximum number of turn is set as 20.}

\subsection{Weakly Supervised Learning Algorithm}
\label{weasul}
Learning with weak supervision is widely used with the rise of data-driven neural
approaches \cite{ratner2020snorkel, mrksic-etal-2017-neural, chang2020unsupervised, bach2017learning, wu2018learning, chang2021jointly}. Our approach incorporates a similar line of work by providing noisy text to a pre-trained model which incorporates  prior knowledge from general-domain text and small in-domain text \cite{peng2020few, chen2019few, harkous2020have} and use it as a weak annotator similar to \citet{ratner2020snorkel}. The primary challenge with the synthetic data is the noise introduced during the generation process, and the noisy labels tend to bring little to no improvement \cite{frenay2013classification}. To train on such noisy data, we employ three step training process: a) pre-training b) generate data with weighted categories c) fine-tuning similar to \citet{chang2021jointly, dehghani2017fidelity}.

\textbf{\textit{Step 1: Pre-train Generation and Quality Modules Jointly}}.
This step involves pre-training the agent with quality modules jointly as explained in Section \ref{tpj}. Quality modules trained on clean data as well as automatically generated negative samples by random sampling. These modules are further fine-tuned on the sampled dialogues from target dialogue corpus at each training iteration. Similarly, we initialized the user also by supervised training on the pre-training dialogue corpus with fine-tuning on target dialogue corpus. (see steps 2-7 of Algorithm \ref{ref:algo}). The fine-tuning steps make use of continual learning to avoid catastrophic forgetting \cite{madotto2020continual, lee2017toward}. 

\textbf{\textit{Step 2: Generates the Weakly Labelled data with Reward categories}}. After the models are initialized with trained parameters, the dialogue simulation 
has been started between the agent and the user (see Section. \ref{diagsim}) to interact with each other and generates the synthetic data with annotated scores with each quality module for every state-action pair in sampled dialogues. During dialogue simulation, we employ Dynamic Blocking mechanism\cite{niu2020unsupervised} to generate novel words and paraphrased responses. Specifically, we generate Top-7 response at each turn and set the agent to exploration for 60 percent of the times and for the rest of the times it exploits by selecting the response from top two ranked responses. We specifically filter the state-action pairs into three reward categories namely, \textit{VeryHigh}, \textit{High} and \textit{Low}. For the state-action pairs whose scores by each module are greater than or equal to 0.8 are put into the \textit{VeryHigh} category. Other, state-action pairs whose scores by each module are between 0.6 and 0.8 are put into the \textit{High} reward category. The rest of all state-action pairs are put into the \textit{Low}
reward category. Additionally, we include state-action pairs sampled from target dialog corpus in Step 1. into the \textit{VeryHigh} category.

\textbf{\textit{Step 3: Update the reward estimator and policy}}. The reward estimator maximizes the log likelihood state-action pairs of higher rewards than the lower ones. The reward estimator $f_{\omega}$, parameterized by $\omega$, and let's say $H$, $V$ and $L$ represents the collection of all state action pairs of \textit{High}, \textit{VeryHigh} and \textit{Low} reward category respectively. 
\begin{multline}
\begin{split}
    \omega^{*} = \argmax \E_{(s_k,a_k) \sim \{H, V\}}[f_{\omega}(s_k, a_k)]\\
    f_{\omega}(s_k, a_k) = \log p_{\omega}(s_k, a_k) = \log \frac{e^{R_{\omega}(s_k, a_k)}}{Z_{\omega}}\\
    R_{\omega}(s_k, a_k) =  \sum_{t=k}^{T} \gamma^{t -k} r_{\omega}(s_t, a_t) \\
    Z_{\omega} = \sum_{\forall (s_k, a_k)} e^{R_{\omega}(s_k, a_k)}
\end{split}
\end{multline}
where $f$ models state-action pairs of H, V and L category as a Boltzmann distribution \cite{takanobu2019guided}. The cost function for reward estimator in terms of trajectories obtained from respective reward categories is given as:
\begin{multline}
        J_f(\omega) = - 0.5*KL ( p_H(s, a) \parallel p_{\omega}(s, a)) \\ 
        - KL(p_{V}(s, a) \parallel p_{\omega}(s, a)) \\
        + KL(p_{L}(s, a) \parallel p_{\omega}(s, a)) 
\end{multline}
It minimize the KL-divergence between reward distribution and the state-action pairs of high and very high reward but maximize the distribution from the ones with low category. The gradient yields:
\begin{multline}
\label{eq:rew}
    \bigtriangledown_{\omega}J_f = 0.5* \E_{(s,a) \sim H}[\bigtriangledown_{\omega}f_{\omega}(s, a)]\\ 
    + \E_{(s,a) \sim V}[\bigtriangledown_{\omega}f_{\omega}(s, a)]
 - \E_{(s, a) \sim L}[\bigtriangledown_{\omega}f_{\omega}(s, a)]
\end{multline}
Since, the dialog policy is required to put the actions atleast to that of high category, i.e. maximize the entropy regularized expected reward ($\E_{\pi}[R] + H(\pi)$) which is effectively minimizes the KL divergence between the policy distribution and Boltzmann distribution.
\begin{multline}
    J_{\pi}(\theta) = - KL(\pi_{\theta}(a|s) \parallel p_{\omega}(s, a)) \\
    = \E_{(s,a) \sim \pi}[f_{\omega}(s, a) - \log \pi_{\theta}(a | s)]\\
    = \E_{(s, a) \sim \pi} [R_{\omega}(s, a)] - \log Z_{\omega} + H(\pi_{\theta})
\end{multline}
where the term $\log Z_{\omega}$ is independent to $\theta$, and
$H(\cdot)$ denotes the entropy of a model. Using likelihood ratio trick the gradient for policy is given as:
\begin{multline}
    \triangledown_{\theta}J_{\pi} = \E_{(s, a) \sim \pi}[(f_{\omega}(s, a) \\
    - \log \pi_{\theta}(a | s ))\triangledown_{\theta} \log \pi_{\theta}(a | s)].
\end{multline}
Hence, the reward is $r_{\omega}(s, a) = f_{\omega}(s, a) - \log \pi_{\theta}(a | s )$ for each state-action pair and the loss function re-written as: 
\begin{multline}
\label{eq:pol}
    J_{\pi}(\theta) = \E_{(s,a) \sim \pi}[\sum_{k=t}^{T}\gamma^{k-t}(f_{\omega}(s_k, a_k) \\
    - \log \pi_{\theta}(a_k | s_k))]
\end{multline}
Like in \citet{takanobu2019guided} the reward estimator $f_{\omega}$ includes the shaping term. Formally, we include next state $s_{t + 1}$ also instead of just $(s_t, a_t)$
\begin{equation}
    f_{\omega}(s_t, a_t, s_{t + 1}) = g_{\omega}(s_t, a_t) + \gamma h(s_{t+1}) - h(s_t) 
\end{equation}
where $h$ is the MLP network with input as pre-sigmoid scores from each quality modules, and $g_{\omega}$ is also the MLP network with input as the concatenation of $E_{CLS}$ as state vector and SBERT sentence embedding of action $a$.

\begin{algorithm}[t]
\caption{Dialogue Policy Learning}
\begin{algorithmic}[1]
\Require Pre-Training corpus $P$, Dialogue Corpus $\mathcal{D}$.
\State Modules $\mathcal{M}$ = \{Semantic Relevance, Semantic Coherence, Consistent Flow\}
\State Do Agent training on $\mathcal{P}$ as in Section \ref{tpj} jointly with modules $\mathcal{M}$
\State User $\mu$ supervised training on $\mathcal{P}$.
\ForEach {\textrm{\textit{training iteration}}} 
    \State Sample dialogues $\mathcal{D}_{H}$ from $\mathcal{D}$ randomly.
    \State Fine-tune user simulator $\mu$ on $\mathcal{D}_{H}$.
    \State Fine-tune Agent and $\mathcal{M}$ on $\mathcal{D}_{H}$ jointly. 
    \State Collect dialog samples $\mathcal{D}_{\pi}$ by executing 
        \StatexIndent[2] the dialog policy $\pi$ and interacting with 
        \StatexIndent[2] $\mu$, $a^{u} \sim \mu( \cdot | s^{u})$, $a \sim \pi( \cdot | s )$ where $s$ 
        \StatexIndent[2] and $s^{u}$ is updated each time after get-  
        \StatexIndent[2] ting response from user and agent re-
        \StatexIndent[2] spectively.
    \State Get weak annotation scores for all $(s, a) \in$  
    \StatexIndent[2] $\mathcal{D_\pi}$ from each of the modules $\mathcal{M}$.
    \State Filtering the $(s, a)$ pairs into \{\textit{VeryHigh}, \StatexIndent[2] \textit{High} and \textit{Low}\} reward categories.
    \State Update the reward estimator $f$ by minimiz-
    \StatexIndent[2] ing $J_{f}$ w.r.t $\omega$ ( Eq.\ref{eq:rew})
    \State Compute reward for each $(s, a) \in \mathcal{D_\pi}$ as,
    $$\hat{r} = f_{\omega}(s_t, a_t, s_{t+1}) - \log \pi( a_t | s_t )$$
    \State Update the policy $\pi_{\theta}$ by minimizing $J_{\pi}$ \StatexIndent[2] w.r.t $\theta$ (Eq. \ref{eq:pol}).
\EndFor
\end{algorithmic}
\label{ref:algo}
\end{algorithm}

\section{Experiments}
We conduct experiments on DailyDialog \cite{li2017dailydialog}, PersonaChat \cite{zhang2018personalizing} and used Gutenberg Dialogue Dataset \cite{csaky2020gutenberg} as a pre-training corpus. We compare our model performance with baselines on various aspects of response quality. 

\subsection{Datasets}
We considered DailyDialog \cite{li2017dailydialog} and PersonaChat \cite{zhang2018personalizing} which are open domain dialog corpus to evaluate our system. DailyDialog contains conversation revolving around various topics pertaining to daily life, and PersonaChat contains conversations between people with their respective persona profiles. These dialogues can be of varying length, we limit the maximum length to 20, that can be fed to the BERT Encoder-Decoder model. Since average length of DailyDialog is 7.9 and that of PersonaChat is 9.4, so most of the dialogues fit easily without truncation from the history. For rest of the dialogues, it can be slided across to include the more recent utterances and remove it from the starting. Since we are mapping the utterances to their corresponding vectors using SBERT, the length of individual utterances truncated automatically and retain only first 512 word pieces in case of longer utterances. For pre-training corpus the vocabulary is limited to 100,000  while the vocabularies for DailyDialog and PersonaChat are  25,000 and 32,768 respectively.

\subsection{Baselines}
We select various multi-turn response generation baselines. The baselines which are not included pre-training are (1) \textbf{HRED}
: Hierarchical encoder-decoder framework \cite{serban2016building} (2) \textbf{VHRED}
: an extension of HRED that generates response with latent variables \cite{10.5555/3298023.3298047} (3) \textbf{HRAN}
: Hierarchical attention mechanism based encoder-decoder framework \cite{xing2018hierarchical} (4) \textbf{ReCoSa}
: Hierarchical transformer based model \cite{zhang2019recosa} (5) \textbf{SSN}: dialogue generation learning
with self-supervision signals extracted from utterance order \cite{wu2019self} (6) \textbf{Transformer-Auxiliary Tasks}: A recent state-of-the are model leaning language generation with joint learning of transformer with auxiliary tasks \cite{zhao2020learning}. The another two baselines from \citet{csaky2020gutenberg} which involve pre-training on the Gutenberg corpus are: (1)\textbf{Transformer}
: 50M parameters version and (2) \textbf{GPT-2}
: Pre-trained model with version of 117M parameters. The repository\footnote{https://github.com/ricsinaruto/gutenberg-dialog} contains these two trained models. 

\subsection{Evaluation Metrics}
We evaluate the performance of our model on various aspects of response quality using both automatic and human evaluation. Although, most of the automatic metrics poorly correlate with human evaluation \cite{liu2016not}, and the recently proposed metrics \cite{li2017adversarial, lowe2017towards, tao2018ruber} are harder to evaluate than perplexity and BLEU \cite{papineni2002bleu}. Additionally, human evaluation has its inherent limitation of bias, cost and replication difficulty \cite{tao2018ruber}. Due to this consensus, some used only automatic metrics \cite{xing2018automatic, xu2018better} and some used only human evaluation \cite{krause2017edina, fang-etal-2018-sounding} while some used both \cite{shen2018nexus, xu2018towards, baheti2018generating, ram2018conversational}.

We mainly used the automatic metrics using the DIALOG-EVAL repository\footnote{https://github.com/ricsinaruto/
dialog-eval}, it contains 17 different metrics, but we measure only a few metrics to facilitate the comparison with the published baselines results. We specifically follow \cite{zhao2020learning} to measure automatic evaluation and human evaluation. For response content quality we measured BLEU-4 \cite{papineni2002bleu} and Perplexity(PPL) \cite{sutskever2014sequence}. Like in \citet{zhao2020learning} used embedding metrics average (AVG), extrema (EXT), and greedy (GRE) measuring similarity between response and
target embedding. Similar to \citet{zhao2020learning} we also measured the informativeness of responses with distinct-1 and distinct-2 that are calculated as the ratios of distinct unigrams and bigrams. 

Since our main objective is not to judge the response quality but to predict the response for long-term success of dialogue. We follow the guidelines as in \citet{li2016deep} to explore both single-turn and multi-turn settings. We picked 500 dialogues from the test set and asked 3 native speakers for their judgement. In the first setting, we asked judges to pick the better response among the one generated by our model and a baseline model (\textbf{Pre-Trained GPT2}) based on various criteria like answerability and semantics. In the second setting, in case of multi-turn we used 200 simulated conversations between RL agent and a user model to judge the whole conversation for responses uttered by agent. In a complete end-to-end conversation we asked the judges to decide which of the simulated conversations are of higher quality. To compare against the RL model we employ baseline model to simulate the 200 conversations with the same starter utterance used by RL model. Automatic and Human evaluation are shown in Table. \ref{tab:auto} and \ref{tab:human} respectively. 

\begin{table*}[]
\resizebox{\textwidth}{!}{%
\begin{tabular}{cllllllll}
\thickhline
\textbf{Dataset} & \textbf{Model} & \textbf{PPL} & \textbf{BLEU} & \textbf{Distinct-1} & \textbf{Distinct-2} & \textbf{Average} & \textbf{Greedy} & \textbf{Extrema} \\ \thickhline

\multirow{7}{*}{DailyDialog} & HRED                             & 56.22 & 0.535 & 1.553 & 3.569 & 81.393 & 65.546 & 48.109 \\
                             & HRAN                             & 47.23 & 0.447 & 1.953 & 7.400 & 83.460 &  67.239 & 49.599 \\
                             & VHRED                            & 44.79 & 0.997 & 1.299  & 6.113  &  83.866  & 67.186 & 48.570 \\
                             & SSN                              & 44.28 & 1.250 & 2.309  & 7.266 & 72.796 & 73.069 & 44.260 \\
                             & ReCoSa                           & 42.34 & 1.121 & 1.987 &  10.180 &  84.763 &  67.557  & 48.957 \\
                             & Transformer-Auxiliary Tasks & 38.60 & 1.658 & 3.457 & 14.954 & 85.224  & 69.518  &  49.069  \\ 
                             & Pre-Trained Transformer                           &  - & 11.5 &  2.92  & 14.7 &  55.1  & 53.5  & 59.8 \\
                                                          
                                                                                       & Pre-Trained GPT2                           &  - & 12.8 &  4.07  & 25.9  &  56.8  & 54.0  & 59.6 \\
                             \cline{2-9}
                             & Our Model                        &  \textbf{20.13} & \textbf{15.171} & \textbf{6.316} & \textbf{28.422} & 85.417 & \textbf{73.118} & \textbf{61.539} \\
                             & Our Model w/o weak supervision                      & 20.51  & 14.718 & 4.611 & 26.752 & \textbf{86.481} & 73.003 & 59.911\\   \hline\hline
\multirow{7}{*}{PersonaChat} & HRED                             & 46.04 & 1.279 & 0.164 &  0.450 & 83.329 & 65.546 & 48.109 \\
                             & HRAN                             & 41.94  & 1.997 & 0.235 &  0.771  & 82.850 &  67.239  & 49.599 \\
                             & VHRED                            & 42.07 &  2.181  & 0.312 &  1.915  &  82.995 &  67.186 &  48.570 \\
                             & SSN                              &  47.90 & 2.288 &  0.637  &  2.623 & 85.002 & 73.069 & 44.260  \\
                             & ReCoSa                           &  34.19 & 2.258 &  0.915 & 4.217 &  83.963 & 67.557 & 48.957 \\
                             & Transformer-Auxiliary Tasks & 33.23 &  2.434 & 1.279 &  5.816 & 83.632 & 69.518 & 49.069 \\ 
                                                          & Pre-Trained Transformer                           & - & 15.5 &  1.04  & 4.8 &  51.3  & 57.5  & 57.1 \\
                                                          
                                                                                       & Pre-Trained GPT2                           & - & 15.3 &  1.82  & 12.9 &  53.6  & 55.9 & 55.8 \\
                             \cline{2-9}
                             & Our Model                        & \textbf{19.78} & \textbf{16.651} & \textbf{2.434} & \textbf{13.912} & 84.941 & \textbf{73.081} & \textbf{59.241} \\ 
                             & Our Model w/o weak supervision                     & 21.49 & 16.017 & 2.318 & 13.274 & \textbf{85.018} & 72.438 & 58.816 \\ \thickhline
\end{tabular}%
}
\caption{Automatic metrics comparison with baselines. Results in bold indicate the best performing model on the corresponding metrics.}
\label{tab:auto}
\vspace{-4mm}
\end{table*}

\begin{table}
\resizebox{\linewidth}{!}{%
\begin{tabular}{clll}
\thickhline
\multicolumn{4}{c}{\textbf{DailyDialog}}                             \\ \hline
\textbf{Setting} & \textbf{RL-Win} & \textbf{RL-Lose} & \textbf{Tie} \\ \hline
Single-Turn general quality    &      0.41            &     0.28              &     0.31         \\
Single-Turn ease to answer     &      0.55         &       0.12                &    0.33          \\
Multi-turn general quality     &       0.76          &     0.13             &      0.11        \\ \hline\hline
\multicolumn{4}{c}{\textbf{PersonaChat}}                             \\ \hline
\textbf{Setting} & \textbf{RL-Win} & \textbf{RL-Lose} & \textbf{Tie} \\ \hline
Single turn general quality    &      0.36       &       0.22           &     0.42          \\
Single-Turn ease to answer     &       0.51            &     0.14              &    0.35          \\
Multi-turn general quality     &      0.71           &       0.17           &      0.12        \\ \thickhline
\end{tabular}%
}
\caption{Human Evaluation Results. Ratios are calculated after taking majority vote among the decisions made by three judges.}
\label{tab:human}
\vspace{-4mm}
\end{table}

\subsection{Results and Discussions}
Table. \ref{tab:auto} reports automatic evaluation metrics on the baseline and the proposed model. Our model outperforms for most of the metrics on both datasets. Since our main idea is to generate the responses for successful conversation in the long run than just evaluating the response quality at each of the turn. This is the main reason of why our model outperforms on both distinct-1 and distinct-2 metrics, in comparison to Transformer-auxiliary task model which also trained jointly with the similar tasks but lacks fine-tuning with the weak supervision signals indicate that an additional training with weakly labelled data improves the generalization performance. Although, we see the perplexity also improves since our model is generating the responses more like humans to optimize the conversation in long run. Similarly, embedding metrics also shown the improvement but little on average since it capturing the sense but due to length mismatch which occurs owing to the fact that our model is generating more novel words with futuristic sense. However, Distinct-\{1,2\} scores shows improvement because of the large pre-trained vocabulary, it gives the model more flexibility to generate novel words without disturbing the sense of the sentence. 

We also note the results for our model without weak supervision training, namely, \textbf{Our Model w/o Weak Supervision}, this model just fine-tunes on the DailyDialog \cite{li2017dailydialog} and PersonaChat \cite{zhang2018personalizing} without generating the weak labelled data. Clearly, the distinct-1 and distinct-2 metrics are lower than the proposed model, because the model tends to generate the repetitive words more frequently. Similarly, the embedding metrics and PPL does not show any improvement over the proposed model except on embedding metric based on Average. However, it performs well on BLEU scores since it learns well to reproduce the responses as in the ground truth but not optimized for a successful conversation in the long run.

Table \ref{tab:auto} also reports the results of another two baselines which are pre-trained models on Gutenberg Dialogue Corpus \cite{csaky2020gutenberg}. These models are fine-tuned on DailyDialog and PersonaChat dataset respectively.  These models although improved much on BLEU scores and distinct-1 and distinct-2 scores since it gets the larger vocab and more enhanced training for learning the language structure. But lags in the embedding metrics indicating the response quality is low. 

Table \ref{tab:human} reports the human evaluation results, the objective for which our model training is to generate the response for a successful conversation in the long run for the multi-turn scenario. Clearly, the evaluation results are up to our expectation, since the RL system does not bring a significant boost in single-turn response quality than the case of multi-turn setting.

\section{Conclusions}
We proposed a weak supervision framework for policy and reward estimation for long-term success of the dialogue by simulating the conversation between a virtual agent and user. Empirical studies on two benchmarks proves the effectiveness of our approach.


\bibliographystyle{acl_natbib}
\bibliography{anthology,acl2021}

\clearpage
\appendix
\section{Implementation Details}
Our implementation uses the open source Huggingface Transformer repository \cite{wolf2020huggingfaces}. Specifically, we have used the base version from sentence transformers pre-trained on millions of paraphrase examples, named as \lq{\textit{paraphrase-distilroberta-base-v1}}\rq. The encoder-decoder framework is initialized with the base version \lq{\textit{bert-base-uncased}}\rq but with configuration of smaller size. The smaller sized model reduces the \lq{\textit{bert-base-uncased}}\rq configuration to 6 transformer layers, has a hidden size of 768, and contains 2 attention heads, \{L=6, H=768, A=2\}. Similar to \citet{gu2020dialogbert} we sum the position embeddings to the output sentence embeddings of size 768 to indicate the user or agent utterances. Odd ones indicate the user utterances and even ones are that of an agent. The MLP network for semantic relevance and semantic coherence used a hidden dimension of 128. The $\Delta$ has been set to best value of 0.54 after performing a grid search in the range of \{0.4, 0.7\} with step size of 0.02. The reward estimator models $g_{\omega}$ using two hidden layers of size 512 and 256 respectively. And, $h$ is modelled using a single hidden layer of size one. In each training iteration the policy and reward estimator are updated with continual learning to avoid catastrophic forgetting mechanism using EWC modified loss, the $\lambda$ value used as a parameter is set to 0.4. Also, at each training iteration the policy and reward parameters are saved if it reduces the perplexity on the validation set (calculated after running for all the batches of the training dataset) and patience is set to 3 as a stopping criterion before we terminate the training.
\end{document}